\documentclass[12pt]{article}
\usepackage{mathbbold}
\usepackage[margin=1.25 in]{geometry}
\begin{document}

{\centering{\Large\bf{Associations among Image Assessments\\ as Cost Functions in Linear Decomposition: MSE, SSIM, and Correlation Coefficient}}

%}

\vspace{0.3 cm} {Jianji Wang, Nanning Zheng, Badong Chen, Jose C. Principe}

{\small{Institute of Artificial Intelligence and Robotics, Xi'an
Jiaotong University, Xi'an, China

wangjianji@mail.xjtu.edu.cn}}

}

{\center{\bf{Abstract}}

}

The traditional methods of image assessment, such as mean squared error (MSE), signal-to-noise ratio (SNR), and Peak signal-to-noise ratio (PSNR), are all based on the absolute error of images. Pearson's inner-product correlation coefficient (PCC) is also usually used to measure the similarity between images. Structural similarity (SSIM) index is another important measurement which has been shown to be more effective in the human vision system (HVS). Although there are many essential differences among these image assessments, some important associations among them as cost functions in linear decomposition are discussed in this paper. Firstly, the selected bases from a basis set for a target vector are the same in the linear decomposition schemes with different cost functions MSE, SSIM, and PCC. Moreover, for a target vector, the ratio of the corresponding affine parameters in the MSE-based linear decomposition scheme and the SSIM-based scheme is a constant, which is just the value of PCC between the target vector and its estimated vector.

\vspace{0.2 cm} {\bf{Key words}}: Image Quality Assessment (IQA), Mean Square Error (MSE), Structural Similarity Index (SSIM), Pearson's Correlation Coefficient (PCC), linear decomposition.

{\center\bf{1. \quad Introduction}

}

In the big data era, there is an increasing importance of images in our lives. We can easily obtain images with cameras in various intelligent devices, and also, from network. Errors usually appear when images are obtained. For example, when we take an image with a camera, the distortion usually happens because of the lens; when an image is downloaded in the internet, errors may also appear for several reasons such as the image transmission and the image compression. In these cases, image quality assessments (IQA) are important tools to measure the effectiveness of different hardware and software systems to preserve image quality. %In fact,

The absolute error-based image assessments, such as the mean square error (MSE),  the signal-to-noise ratio (SNR), and the Peak signal-to-noise ratio (PSNR), are the most common measurements to measure image error. Intuitively, the addition of the pixel-by-pixel squared errors between two images is the square of the distance between them. MSE is the average squared distance for the corresponding pixel in two images. For two image blocks ${\bf{x}}$ and ${\bf{y}}$, if the pixels in ${\bf{x}}$ are $x_1,x_2,\cdots,x_p$, and the pixels in ${\bf{y}}$ are $y_1,y_2,\cdots,y_p$, then MSE value between ${\bf{x}}$ and ${\bf{y}}$ can be calculated as following,
\begin{equation}
{\rm{MSE}}({\bf{x}},{\bf{y}}) = \frac{1}{p}\sum_{i=1}^{p}{(y_i-x_i)^2}.
\end{equation}
Therefore, MSE is a pixel-based error measurement. SNR and PSNR are both derived from MSE. These absolute error-based assessments are not only used to measure image quality, but also used to measure almost all kinds of signals.

Pearson's inner-product correlation coefficient (PCC) is an important statistic which is used to measure the correlation between two vectors. For non-zero variance image blocks ${\bf{x}}$ and ${\bf{y}}$ discussed above, if $\sigma_{\bf{x}}$ and $\sigma_{\bf{y}}$ are the standard deviations of the pixels in the image blocks ${\bf{x}}$ and ${\bf{y}}$, respectively,  $\overline{x}$ and $\overline{y}$ are the means of the pixels in the image blocks ${\bf{x}}$ and ${\bf{y}}$, respectively, $\theta$ is the angle between ${\bf{x}}$ and ${\bf{y}}$ when we take them as two column vectors, and $\sigma_{\bf{xy}}$ is the covariance between ${\bf{x}}$ and ${\bf{y}}$, correlation coefficient $r_{\bf{xy}}$ between ${\bf{x}}$ and ${\bf{y}}$ then can be calculated as following,
\begin{equation}
\begin{array} {lcl}
{r_{\bf{xy}}} = \frac{\sum_{i=1}^{p}(x_i-\overline{x})(y_i-\overline{y})}{\sqrt{\sum_{i=1}^{p}(x_i-\overline{x})^2}\sqrt{\sum_{i=1}^{p}(y_i-\overline{y})}} \\
= \frac{\sigma_{\bf{xy}}}{\sigma_{\bf{x}}\sigma_{\bf{y}}}=\cos\theta
\end{array}.
\end{equation}
Correlation coefficient is also usually used to measure the similarity between images \cite{Xue,SPL1}. Essentially, correlation coefficient measures the correlation of the structure of two signals. Hence, correlation coefficient is actually a correlation-based assessment.

Structural similarity index (SSIM), proposed by Wang and Bovik \cite{Wang1}, aims to improve the effectiveness of IQA in human visual systems (HVS). In SSIM, the errors are taken as three parts: the luminance error, the contrast error, and the structure error. By combining the three parts, SSIM gets a form as
\[{\rm{SSIM}}({\bf{x}},{\bf{y}}) = \left[ {\frac{{2{\mu _{\bf{x}}}{\mu _{\bf{y}}} + {\varepsilon _1}}}{{\mu _{\bf{x}}^2 + \mu _{\bf{y}}^2 + {\varepsilon _1}}}} \right]\left[ {\frac{{2{\sigma _{\bf{x}}}{\sigma _{\bf{y}}} + {\varepsilon _2}}}{{\sigma _{\bf{x}}^2 + \sigma _{\bf{y}}^2 + {\varepsilon _2}}}} \right],\]
where ${\mu _{\bf{x}}} = \overline{x}$, ${\mu _{\bf{y}}} = \overline{y}$, and ${\varepsilon _1},{\varepsilon _2} <\!< 1$ are two small positive constants. If the variance of a given image block ${\bf{y}}$ is zero, ${\bf{y}}$ can be losslessly linearly expressed by ${\bf{1}}$ with all ones. Because this paper focuses on linear decomposition, here we only consider the image blocks with non-zero variance. At this case, we can set ${\varepsilon _1}={\varepsilon _2}=0$, then SSIM gets a simpler form
\begin{equation}
{\rm{SSIM}}({\bf{x}},{\bf{y}}) = \frac{4\mu_{\bf{x}}\mu_{\bf{y}}\sigma_{\bf{xy}}}{(\mu_{\bf{x}}^2+\mu_{\bf{y}}^2)(\sigma_{\bf{x}}^2+\sigma_{\bf{y}}^2)}.
\end{equation}

SSIM has experienced a fast development from the baseline SSIM to various forms \cite{Wang2,Wang3,notHVS}. Although many researchers do not think that SSIM do much to improve the effectiveness of image quality assessment in HVS \cite{notHVS,notgood}, SSIM has been widely accepted.

As described above, MSE is a pixel-based image assessment, correlation coefficient is a correlation-based image assessment, and SSIM is a structure-based image assessment for HVS. Thus, there are many significant and essential differences among them. In this paper, we do not focus on these important differences among the image quality assessments. On the contrary, some interesting associations are disccussed when we take the image quality assessments as the cost functions in linear decomposition.

{\center\bf{2. \quad Linear Decomposition}

}

Linear decomposition plays an important role in various fields such as linear approximation, sparse coding \cite{SC1}, and portfolio \cite{portfolio}. Especially, sparse coding is an important tool in image processing. Suppose we have a vector set $\bf{X}$ with $n$ vectors \{${\bf{x}}_1, {\bf{x}}_2,$ $ \cdots, {\bf{x}}_n$\}, and each vector is an image block with size $l\times l$, $p=l\times l$. For an image block $\bf{y}$ with size $l\times l$, we need to find a linear transformation ${\bf{x}}=s_1{\bf{x}}_1+s_2{\bf{x}}_2+\cdots+s_n{\bf{x}}_n+o\bf{1}$ to linearly approximate $\bf{y}$, where $s_1, s_2, \cdots, s_n$ and $o$ are the linear coefficients and $\bf{1}$ is the block with all ones. In linear approximation, the vector set $\bf{X}$ is called codebook, and each vector in the codebook is called codeword; In sparse coding, $\bf{X}$ is called basis set and a codeword is called a basis. $\bf{X}$ can be trained by K\_SVD algorithm \cite{K_SVD}.  Here, we use the names in sparse coding.

Suppose that the image block $\bf{y}$ is linear decomposed with $m$ bases in the basis set, then linear decomposition is the optimization problem as following:
\begin{equation}
\begin{array}{l}
\mathop {\min }\limits_{{{\bf{x}}_{{i_1}}},{{\bf{x}}_{{i_2}}}, \cdots ,{{\bf{x}}_{{i_m}}} \in \{ {{\bf{x}}_1},{{\bf{x}}_2}, \cdots ,{{\bf{x}}_n}\} } d({\bf{y}},{\bf{x}})\\
{\bf{x}} = {s_{{i_1}}}{{\bf{x}}_{{i_1}}} + {s_{{i_2}}}{{\bf{x}}_{{i_2}}} +  \cdots  + {s_{{i_m}}}{{\bf{x}}_{{i_m}}} + o{\bf{1}}
\end{array},
\end{equation}
where $d({\bf{y}},{\bf{x}})$ is the distance of $\bf{x}$ and $\bf{y}$ under a cost function. The traditional linear decomposition scheme takes MSE between ${\bf{x}}$ and ${\bf{y}}$ as the cost function, and we will discuss different linear decomposition schemes with different cost functions in this paper.

Strictly speaking, Pearson's correlation coefficient and the SSIM index are not distance functions because both of them do not satisfy the triangle inequality of distance function. However, it is meaningful that we take the absolute value of correlation coefficient and the absolute value of the SSIM index as the cost functions in image quality assessment. Moreover, the results we obtain in this paper shows these cost functions are equivalent with MSE in selecting basis vectors from the same basis set for an given image block to be encoded, which further proves the validity of taking the absolute values of correlation coefficient and the SSIM index as the cost functions in image quality assessment.

Many methods had been proposed to search the corresponding basis vectors with non-zero coefficients from the basis set $\bf{X}$ for a target vector $\bf{y}$. Matching Pursuit (MP) \cite{MP} and Orthogonal Matching Pursuit (OMP) \cite{OMP} are two common used technologies. MP is an iterative method which selects a basis vector from the basis set in its each step. OMP is an improved technology of MP with better convergence.

In linear approximation, if only one $s_i$ in $s_1, s_2, \cdots, s_n$ is non-zero, $i=1,2,\cdots,n$, then linear approximation degrades to be vector quantization (VQ). Some works had been performed on the MSE-based VQ scheme and the SSIM-based VQ scheme \cite{My,My2,Other1}. Several works also discussed the application of SSIM to linear approximation \cite{FormofSSIM,restoration,Maths}. Here we will discuss the associations among the different linear decomposition schemes based on different cost functions Pearson's correlation coefficient, MSE, and SSIM.

{\center\bf{3. \quad Linear Decomposition with Different Cost Functions}

}

According to linear decomposition, a linear transformation $s_1{\bf{x}}_1+s_2{\bf{x}}_2+\cdots+s_n{\bf{x}}_n+o\bf{1}$ with a few non-zero $s_i$ needs to be found to approximate a target vector $\bf{y}$, $i=1,2,\cdots,n$.

Without loss of generality, assume ${\bf{x}}_1, {\bf{x}}_2, \cdots, {\bf{x}}_m$ are the selected basis vectors with non-zero value of $s_i$, and ${\bf{x}}=s_1{\bf{x}}_1+s_2{\bf{x}}_2+\cdots+s_m{\bf{x}}_m+o\bf{1}$ is the best linear approximation for the target vector $\bf{y}$. Let the standard deviation of ${\bf{x}}_i$ is $\sigma_i$, the standard deviations of ${\bf{x}}$ and ${\bf{y}}$ are $\sigma_{\bf{x}}$ and $\sigma_{\bf{y}}$, respectively, the means of ${\bf{x}}_i$ and ${\bf{y}}$ are $\mu_i$ and $\mu_{\bf{y}}$, respectively, the covariance between ${\bf{x}}_i$ and ${\bf{x}}_j$ is $\sigma_{ij}$, the covariance between ${\bf{x}}_i$ and ${\bf{y}}$ is $\sigma_{i{\bf{y}}}$, and the covariance between ${\bf{x}}$ and ${\bf{y}}$ is $\sigma_{\bf{xy}}$, $i=1,2,\cdots,n$.

Let %$\bbsigma$ is the covariance matrix of ${\bf{x}}_1, {\bf{x}}_2, \cdots, {\bf{x}}_m$,\\
$\bbsigma  = \left[ {\begin{array}{*{20}{c}}
{{\sigma _{11}}}&{{\sigma _{12}}}& \cdots &{{\sigma _{1m}}}\\
{{\sigma _{21}}}&{{\sigma _{22}}}& \cdots &{{\sigma _{2m}}}\\
 \vdots & \vdots & \ddots & \vdots \\
{{\sigma _{m1}}}&{{\sigma _{m2}}}& \cdots &{{\sigma _{mm}}}
\end{array}} \right]$, ${\bf{s}} = \left[ {\begin{array}{*{20}{c}}
{{s_1}}\\
{{s_2}}\\
 \vdots \\
{{s_m}}
\end{array}} \right]$, and ${\bbsigma _{ \bullet {\bf{y}}}} = {\left[ {\begin{array}{*{20}{c}}
{{\sigma _{1{\bf{y}}}}}&{{\sigma _{2{\bf{y}}}}}& \cdots &{{\sigma _{m{\bf{y}}}}}
\end{array}} \right]^{\rm{T}}}$.\

Then the different linear decomposition schemes with different cost functions will be discussed below.

\vspace{0.1cm} \noindent{\bf{3.1 \ Correlation Coefficient}}\vspace{0.1cm}

Although the parameters $s_1, s_2, \cdots, s_m$ in the linear transformation cannot be calculated by taking Pearson's inner-product correlation coefficient as the cost function, we can still take correlation coefficient as a cost function to decide which basis vector should be chosen for the target vector in different steps of MP searching or OMP searching.

As it is known, correlation coefficient gets the values -$1$ or 1 if two vectors are linear dependent, and it gets the value zero if two vectors are perpendicular with each other. In the above linear transformation, we can set $s_i$ as any negative values or positive values, $i=1,2,\cdots,n$, so here we can only consider the absolute value of Pearson's correlation coefficient, which can measure the strength of correlation for two vectors.

According to the analysis above, for a given image block $\bf{y}$, we only need to maximize $\left| {{r_{{\bf{xy}}}}} \right|$ to find the best ${\bf{x}}$ for ${\bf{y}}$ when we take the correlation coefficient as the cost function.

\vspace{0.1cm} \noindent{\bf{3.2 \ Mean Square Error}}\vspace{0.1cm}

The MSE-based linear decomposition scheme is the traditional linear decomposition.  Here we firstly analyze the relationship between ${\bf{x}}$ and ${\bf{y}}$ under the cost function of MSE. We have
\[\begin{array}{l}
{\rm{MSE}}({\bf{x}},{\bf{y}}) = \frac{1}{p}\left\| {{\bf{y}} - {\bf{x}}} \right\|_2^2\\
 = \frac{1}{p}\left\| {{\bf{y}} - ({s_1}{{\bf{x}}_1} + {s_2}{{\bf{x}}_2} +  \cdots  + {s_m}{{\bf{x}}_m} + o{\bf{1}})} \right\|_2^2
\end{array}.\]

To minimize ${\rm{MSE}}({\bf{x}},{\bf{y}})$, we need to solve the below equations
\begin{equation}
\begin{array}{l}
\frac{{\partial {\rm{MSE}}({\bf{x}},{\bf{y}})}}{{\partial o}} = 0\\
\frac{{\partial {\rm{MSE}}({\bf{x}},{\bf{y}})}}{{\partial {s_i}}} = 0
\end{array}.
\end{equation}

From the first equation in Eq. (5), we can obtain
\begin{equation}
o = {\mu _{\bf{y}}} - {s_1}{\mu _1} - {s_2}{\mu _2} -  \cdots  - {s_m}{\mu _m}.
\end{equation}

Substitute Eq. (6) into ${\rm{MSE}}({\bf{x}},{\bf{y}})$, then we have
\begin{equation}
\begin{array}{l}
{\rm{MSE}}({\bf{x}},{\bf{y}}) = \frac{1}{p}\left\| {{\bf{y}} - {\bf{x}}} \right\|_2^2\\
 = \frac{1}{p}(\sigma _{\bf{y}}^2 - 2{\sigma _{{\bf{xy}}}} + \sigma _{\bf{x}}^2)\\
 = \frac{1}{p}(\sigma _{\bf{y}}^2 - 2\sum\limits_i {{s_i}{\sigma _{i{\bf{y}}}}}  + \sum\limits_i {\sum\limits_j {{s_i}{s_j}{\sigma _{ij}}} } )
\end{array}.
\end{equation}

According to Eq. (7) and the second equation in Eq. (5), we have
\begin{equation}
{\sigma _{i{\bf{y}}}} = \sum\limits_j {{s_j}{\sigma _{ij}}}.
\end{equation}

From Eq. (8) we can calculate all the $s_i$ for ${\bf{x}}$, and we can also obtain
\begin{equation}
{\sigma _{{\bf{xy}}}} = \sum\limits_i {{s_i}} {\sigma _{i{\bf{y}}}} = \sum\limits_i {{s_i}} \sum\limits_j {{s_j}{\sigma _{ij}}}  = \sigma _{\bf{x}}^2.
\end{equation}

It shows in Eq. (9) that the variance of the best ${\bf{x}}$ chosen for ${\bf{y}}$ in the MSE-based scheme is kept the same with the covariance between ${\bf{x}}$ and ${\bf{y}}$. Thus, according to Eq. (7),
\begin{equation}
\begin{array}{l}
{\rm{MSE}}({\bf{x}},{\bf{y}}) = \frac{1}{p}(\sigma _{\bf{y}}^2 - {\sigma _{{\bf{xy}}}})\\
 = \frac{{\sigma _{\bf{y}}^2}}{p}(1 - \frac{{{\sigma _{{\bf{xy}}}}}}{{\sigma _{\bf{y}}^2}}\frac{{{\sigma _{{\bf{xy}}}}}}{{\sigma _{\bf{x}}^2}}) = \frac{{\sigma _{\bf{y}}^2}}{p}(1 - r_{{\bf{xy}}}^2)
\end{array}.
\end{equation}

Hence, for a given image block ${\bf{y}}$ to be linear decomposed, we have
\begin{equation}
\min {\rm{MSE}}({\bf{x}},{\bf{y}}) \Leftrightarrow \max \left| {{r_{{\bf{xy}}}}}\right|.
\label{result1}
\end{equation}

This interesting result shows that the linear transformations ${\bf{x}}$ chosen for ${\bf{y}}$ in the correlation coefficient-based linear decomposition scheme and the MSE-based scheme both have the maximum absolute value of ${r_{\bf{xy}}}$ in their own schemes.

From Eq. (8), we have
\begin{equation}{\bf{s}} = {\bbsigma ^{ - 1}}\bbsigma _{ \bullet {\bf{y}}}\end{equation}
and
\begin{equation}{\sigma _{{\bf{xy}}}} = \bbsigma _{ \bullet {\bf{y}}}^{\rm{T}}{\bbsigma ^{ - 1}}{\bbsigma _{ \bullet {\bf{y}}}}.\end{equation}

Hence, according to Eq. (10), for a given image block ${\bf{y}}$ to be encoded, we have
\begin{equation}\min {\rm{MSE}}({\bf{x}},{\bf{y}}) \Leftrightarrow \max {\sigma _{{\bf{xy}}}} = \max (\bbsigma _{ \bullet {\bf{y}}}^{\rm{T}}{\bbsigma ^{ - 1}}{\bbsigma _{ \bullet {\bf{y}}}}).\end{equation}

\vspace{0.1cm} \noindent{\bf{3.3 \ Structural Similarity Index}}\vspace{0.1cm}

Here we discuss the structural similarity (SSIM) index-based linear decomposition scheme.

The value of SSIM index belongs to [-1,1]. When the value of SSIM is near to 1, two images are almost the same; when the value of SSIM is -1, two images have the same values of mean and the addition of the corresponding pixel in the two images equals double of the mean of them. Hence, both the values of 1 and -1 are the targets for linear decomposition. To maximize or minimize the value of SSIM index, we need to solve the equations,
\begin{equation}
\begin{array}{l}
\frac{{\partial {\rm{SSIM}}({\bf{x}},{\bf{y}})}}{{\partial o}} = 0\\
\frac{{\partial {\rm{SSIM}}({\bf{x}},{\bf{y}})}}{{\partial {s_i}}} = 0
\end{array}.
\end{equation}

According to the first equation in Eq. (15), we have the same result of Eq. (6) for the coefficient $o$. Then we substitute Eq. (6) into the expression of SSIM index, we have
\begin{equation}
{\rm{SSIM}}({\bf{x}},{\bf{y}})= \frac{{2{\sigma _{{\bf{xy}}}}}}{{\sigma _{\bf{x}}^2 + \sigma _{\bf{y}}^2}} = \frac{{2\sum\limits_i {{s_i}} {\sigma _{i{\bf{y}}}}}}{{\sum\limits_i {{s_i}} \sum\limits_j {{s_j}{\sigma _{ij}}}  + \sigma _{\bf{y}}^2}}.
\end{equation}

Solve the second equation in Eq. (15), we obtain
\[{\sigma _{i{\bf{y}}}}(\sum\limits_i {{s_i}} \sum\limits_j {{s_j}{\sigma _{ij}}}  + \sigma _{\bf{y}}^2) = 2\sum\limits_i {{s_i}} {\sigma _{i{\bf{y}}}}\sum\limits_j {{s_j}{\sigma _{ij}}} . \]

Hence,
\begin{equation}
{\sigma _{i{\bf{y}}}}(\sigma _{\bf{x}}^2 + \sigma _{\bf{y}}^2) = 2{\sigma _{{\bf{xy}}}}\sum\limits_j {{s_j}{\sigma _{ij}}}.
\end{equation}

From Eq. (17),
\[(\sigma _{\bf{x}}^2 + \sigma _{\bf{y}}^2)\sum\limits_i {{s_i}} {\sigma _{i{\bf{y}}}} = 2{\sigma _{{\bf{xy}}}}\sum\limits_i {{s_i}} \sum\limits_j {{s_j}{\sigma _{ij}}}. \]

Because
\[\begin{array}{l}
{\sigma _{{\bf{xy}}}} = \sum\limits_i {{s_i}} {\sigma _{i{\bf{y}}}}\\
\sigma _{\bf{x}}^2=\sum\limits_i {{s_i}} \sum\limits_j {{s_j}{\sigma _{ij}}} \end{array},
\]
we have
\begin{equation}\sigma _{\bf{y}}^2 = \sigma _{\bf{x}}^2.\end{equation}

As Eq. (18) shows, the variance of the best ${\bf{x}}$ chosen for ${\bf{y}}$ in the SSIM-based linear decomposition scheme is kept the same as the variance of ${\bf{y}}$.

Lastly, we have
\[ {{\rm{SSIM}}({\bf{x}},{\bf{y}})} = {\frac{{2{\sigma _{{\bf{xy}}}}}}{{\sigma _{\bf{x}}^2 + \sigma _{\bf{y}}^2}}} = {\frac{{{\sigma _{{\bf{xy}}}}}}{{\sigma _{\bf{x}}^2}}} = {\frac{{{\sigma _{{\bf{xy}}}}}}{{{\sigma _{\bf{x}}}{\sigma _{\bf{y}}}}}} = {{r_{{\bf{xy}}}}}.\]

Because both maximization and minimization of SSIM is equivalent to maximizing the absolute value of SSIM, we have
\begin{equation}\max \left| {{\rm{SSIM}}({\bf{x}},{\bf{y}})} \right|\Leftrightarrow \max \left| {{r_{{\bf{xy}}}}} \right|.\end{equation}

From these results, we get the first association as following:

{\bf{Association 1}.}  Maximizing ${\left| {{r_{{\bf{xy}}}}} \right|}$ in the correlation coefficient-based linear decomposition scheme, minimizing ${{\rm{MSE}}({\bf{x}},{\bf{y}})}$ in the MSE-based linear decomposition scheme, and maximizing ${\left|{\rm{SSIM}}({\bf{x}},{\bf{y}})\right|}$ in the SSIM-based linear decomposition scheme are all equivalent to maximizing ${\left| {{r_{{\bf{xy}}}}} \right|}$ in their own schemes.

According to Eq. (17) and Eq. (18), we have
\begin{equation}
\begin{array}{l}
{\bf{s}} = \frac{{\sigma _{\bf{y}}^2}}{{{\sigma _{{\bf{xy}}}}}}{\bbsigma ^{ - 1}}{\bbsigma _{ \bullet {\bf{y}}}}\\
{\sigma _{{\bf{xy}}}} = {{\bf{s}}^{\rm{T}}}{\bbsigma _{ \bullet {\bf{y}}}}
\end{array}.
\end{equation}

Thus,
\[{\bf{s}}{{\bf{s}}^{\rm{T}}}{\bbsigma _{ \bullet {\bf{y}}}} = \sigma _{\bf{y}}^2{\bbsigma ^{ - 1}}{\bbsigma _{ \bullet {\bf{y}}}}.\]

Make a premultiplication of $\bbsigma _{ \bullet {\bf{y}}}$ in both sides, we have
\begin{equation}\sigma _{{\bf{xy}}}^2 = \sigma _{\bf{y}}^2\bbsigma _{ \bullet {\bf{y}}}^{\rm{T}}{\bbsigma ^{ - 1}}{\bbsigma _{ \bullet {\bf{y}}}}.\end{equation}

Hence, for a given target vector ${\bf{y}}$, according to Eq. (16) and Eq. (18), we obtain
\begin{equation}
\max {\rm{SSIM}}({\bf{x}},{\bf{y}}) \Leftrightarrow \max \left| {{\sigma _{{\bf{xy}}}}} \right| \Leftrightarrow \max (\bbsigma _{ \bullet {\bf{y}}}^{\rm{T}}{\bbsigma ^{ - 1}}{\bbsigma _{ \bullet {\bf{y}}}}).
\end{equation}

{\center\bf{4. \quad The associations}

}

For comparison of different schemes, we consider the case of the same number of basis vectors to linearly approximate the target vector ${\bf{y}}$.

Although the target functions in the different linear decomposition schemes are all equivalent to maximizing the absolute value of correlation coefficient, we cannot confirm these schemes are totally same. For example, for a given image block ${\bf{y}}$ to be encoded, according to Eq. (9), the variance of the final approximation ${\bf{x}}$ equals to the covariance between ${\bf{x}}$ and ${\bf{y}}$ in the MSE-based linear decomposition scheme, but according to Eq. (18) the variance of ${\bf{x}}$ equals to the variance of ${\bf{y}}$ in the SSIM-based scheme. Moreover, it is possible that the affine parameters, the selected basis vectors for the given target vector, and the value of ${r_{{\bf{x^{\rm{MSE}}y}}}}$ and ${r_{\bf{x^{\rm{SSIM}}y}}}$ are different in different schemes.

\vspace{0.1cm} \noindent{\bf{4.1 \ The Selected Basis Vectors}}\vspace{0.1cm}

There are no affine parameters in the expression $\bbsigma _{ \bullet {\bf{y}}}^{\rm{T}}{\bbsigma ^{ - 1}}{\bbsigma _{ \bullet {\bf{y}}}}$, which is only relevant with ${\bf{x}}_1$, ${\bf{x}}_2$, $\cdots$, ${\bf{x}}_m$, and ${\bf{y}}$. Therefore, according to Eq. (14) and Eq. (22), we have the second association of the schemes as following:

{\bf{Association 2}.}  The selected basis vectors from the same basis set for the target vector ${\bf{y}}$ in the the MSE-based linear decomposition scheme and in the SSIM-based linear decomposition schemes are totally the same.

\vspace{0.1cm} \noindent{\bf{4.2 \ The Cost Functions}}\vspace{0.1cm}

Assume the best linear approximations of ${\bf{y}}$ are ${\bf{x}}^{\rm{MSE}}$ and ${\bf{x}}^{\rm{SSIM}}$ in the MSE-based  linear decomposition scheme and the SSIM-based linear decomposition scheme, respectively.

According to Eq. (9) and Eq. (13), we have
\begin{equation}
\left|{r_{{{\bf{x}}^{{\rm{MSE}}}}{\bf{y}}}}\right| = \frac{{{\sigma _{{{\bf{x}}^{{\rm{MSE}}}}{\bf{y}}}}}}{{{\sigma _{{{\bf{x}}^{{\rm{MSE}}}}}}{\sigma _{\bf{y}}}}} = \frac{{\sqrt {{\sigma _{{{\bf{x}}^{{\rm{MSE}}}}{\bf{y}}}}} }}{{{\sigma _{\bf{y}}}}} = \frac{{\sqrt {\bbsigma _{ \bullet {\bf{y}}}^{\rm{T}}{\bbsigma ^{ - 1}}{\bbsigma _{ \bullet {\bf{y}}}}} }}{{{\sigma _{\bf{y}}}}}.
\end{equation}

According to Eq. (18) and Eq. (21), we can also obtain
\begin{equation}
\left|{r_{{{\bf{x}}^{{\rm{SSIM}}}}{\bf{y}}}}\right| = \frac{\left|{{\sigma _{{{\bf{x}}^{{\rm{SSIM}}}}{\bf{y}}}}}\right|}{{{\sigma _{{{\bf{x}}^{{\rm{SSIM}}}}}}{\sigma _{\bf{y}}}}} = \frac{\left|{{\sigma _{{{\bf{x}}^{{\rm{SSIM}}}}{\bf{y}}}}}\right|}{{\sigma _{\bf{y}}^2}} = \frac{{\sqrt {\bbsigma _{ \bullet {\bf{y}}}^{\rm{T}}{\bbsigma ^{ - 1}}{\bbsigma _{ \bullet {\bf{y}}}}} }}{{{\sigma _{\bf{y}}}}}.
\end{equation}

Thus,
\[\left| {{r_{{{\bf{x}}^{{\rm{MSE}}}}{\bf{y}}}}} \right| =  \left| {{r_{{{\bf{x}}^{{\rm{SSIM}}}}{\bf{y}}}}}\right|. \]

Then we have the third association:

{\bf{Association 3}.}  Minimizing ${{\rm{MSE}}({\bf{x}},{\bf{y}})}$ in the MSE-based linear decomposition scheme is equivalent to maximizing ${\left| {\rm{SSIM}}({\bf{x}},{\bf{y}})\right|}$ in the SSIM-based linear decomposition scheme, and they are both equivalent to maximizing ${\left| {{r_{{\bf{xy}}}}} \right|}$.

\vspace{0.1cm} \noindent{\bf{4.3 \ Affine Parameters}}\vspace{0.1cm}

Assume the linear parameters are $s_1^{{\rm{MSE}}}$, $s_2^{{\rm{MSE}}}, \cdots$, $s_m^{{\rm{MSE}}}$, and $o^{{\rm{MSE}}}$ in the MSE-based linear decomposition scheme, and the linear parameters are $s_1^{{\rm{SSIM}}}$, $s_2^{{\rm{SSIM}}}, \cdots$, $s_m^{{\rm{SSIM}}}$, and $o^{{\rm{SSIM}}}$ in the SSIM-based linear decomposition scheme. We have
%\begin{equation}
\[\begin{array}{l}
{{\bf{x}}^{{\rm{MSE}}}} = s_1^{{\rm{MSE}}}{\bf{x}}_1 + s_2^{{\rm{MSE}}}{\bf{x}}_2 +  \cdots  + s_m^{{\rm{MSE}}}{\bf{x}}_m
 + {o^{{\rm{MSE}}}}{\bf{1}}\\
{{\bf{x}}^{{\rm{SSIM}}}} = s_1^{{\rm{SSIM}}}{\bf{x}}_1 + s_2^{{\rm{SSIM}}}{\bf{x}}_2 +  \cdots  + s_m^{{\rm{SSIM}}}{\bf{x}}_m
 + {o^{{\rm{MSE}}}}{\bf{1}}
\end{array}.\]
%\end{equation}

Eqs. (8) and (17) offer the method to calculate affine parameters in the MSE-based scheme and the SSIM-based scheme, respectively. We list them below for comparison:
\begin{equation}
\begin{array}{l}
{\sigma _{i{\bf{y}}}} = \sum\limits_j {s_j^{{\rm{MSE}}}{\sigma _{ij}}} \\
{\sigma _{i{\bf{y}}}} = \sum\limits_j {{r_{{\bf{xy}}}}s_j^{{\rm{SSIM}}}{\sigma _{ij}}}
\end{array}.
\end{equation}

The second equation set in Eq. (25) is transformed from Eq. (17) by use of Eq. (18).

Because ${\sigma _{i{\bf{y}}}}$, ${{\sigma _{ij}}}$ are the same in both equation sets in Eq. (25), we have
\begin{equation}
\frac{{s_i^{{\rm{MSE}}}}}{{s_i^{{\rm{SSIM}}}}} = {r_{{\bf{xy}}}}.
\end{equation}

This equation offers the fourth association between different schemes:

{\bf{Association 4}.}  Although the affine parameters are different in the MSE-based scheme and the SSIM-based scheme, the ratio of the affine parameters in the MSE-based scheme and the corresponding affine parameters in the SSIM-based scheme is a constant, and the constant is just the correlation coefficient between ${\bf{x}}$ and ${\bf{y}}$.

By using Associations 2 and 4, the data storage in sparse coding can be optimized. Because data storage is beyond the scope of this paper, we will discuss it in another paper.

{\center\bf{5. \quad Conclusion}

}

Although there are many essential differences among image assessments MSE, SSIM, and Pearson's correlation coefficient, in this work we have shown several interesting and important associations among them when they are used as the cost functions in linear decomposition. Firstly, the minimization of the MSE value between the target vector and its estimated vector in the MSE-based linear decomposition scheme is equivalent to maximizing the absolute value of SSIM in the SSIM-based linear decomposition scheme, and they are both equivalent to maximizing the correlation coefficient between them. Secondly, the selected basis vectors from the same basis set for a given target vector are the same in both the linear decomposition schemes with the cost functions MSE and SSIM. Moreover, the ratio of the affine parameters in the MSE-based scheme and the corresponding affine parameters in the SSIM-based scheme is a constant, and the constant is just the correlation coefficient between the target vector and its estimated vector. By using these associations, data storage of sparse coding can be optimized.

\end{document}